\documentclass[10pt,twocolumn,letterpaper]{article}

\usepackage{iccv}
\usepackage{times}
\usepackage{epsfig}
\usepackage{graphicx}
\usepackage{amsmath}
\usepackage{amssymb}

\usepackage{color}
\usepackage{subfig}
\usepackage{soul}
\usepackage{multirow}
\usepackage{bbm}
\usepackage[dvipsnames]{xcolor}

\usepackage[pagebackref=true,breaklinks=true,letterpaper=true,colorlinks,bookmarks=false]{hyperref}

\iccvfinalcopy 

\begin{document}

\title{Robustified Domain Adaptation}

\author{%
Jiajin Zhang$^*$ \quad Hanqing Chao\thanks{J.Z. and H.C. are co-first authors.} \quad Pingkun Yan\thanks{Corresponding author.}\\
Deep Imaging Analytics Lab (DIAL) \\ Rensselaer Polytechnic Institute \\ \{zhangj41, chaoh, yanp2\}@rpi.edu}

\maketitle

\begin{abstract}
Unsupervised domain adaptation (UDA) is widely used to transfer knowledge from a labeled source domain to an unlabeled target domain with different data distribution. While extensive studies attested that deep learning models are vulnerable to adversarial attacks, the adversarial robustness of models in domain adaptation application has largely been overlooked. 
This paper points out that the inevitable domain distribution deviation in UDA is a critical barrier to model robustness on the target domain. To address the problem, we propose a novel Class-consistent Unsupervised Robust Domain Adaptation (CURDA) framework for training robust UDA models. With the introduced contrastive robust training and source anchored adversarial contrastive losses, our proposed CURDA framework can effectively robustify UDA models by simultaneously minimizing the data distribution deviation and the distance between target domain clean-adversarial pairs without creating classification confusion. Experiments on several public benchmarks show that CURDA can significantly improve model robustness in the target domain with only minor cost of accuracy on the clean samples. 
\end{abstract}

\section{Introduction}
\label{Sec:intro}

\begin{figure}[t]
\centering
{\includegraphics[width=1.\linewidth]{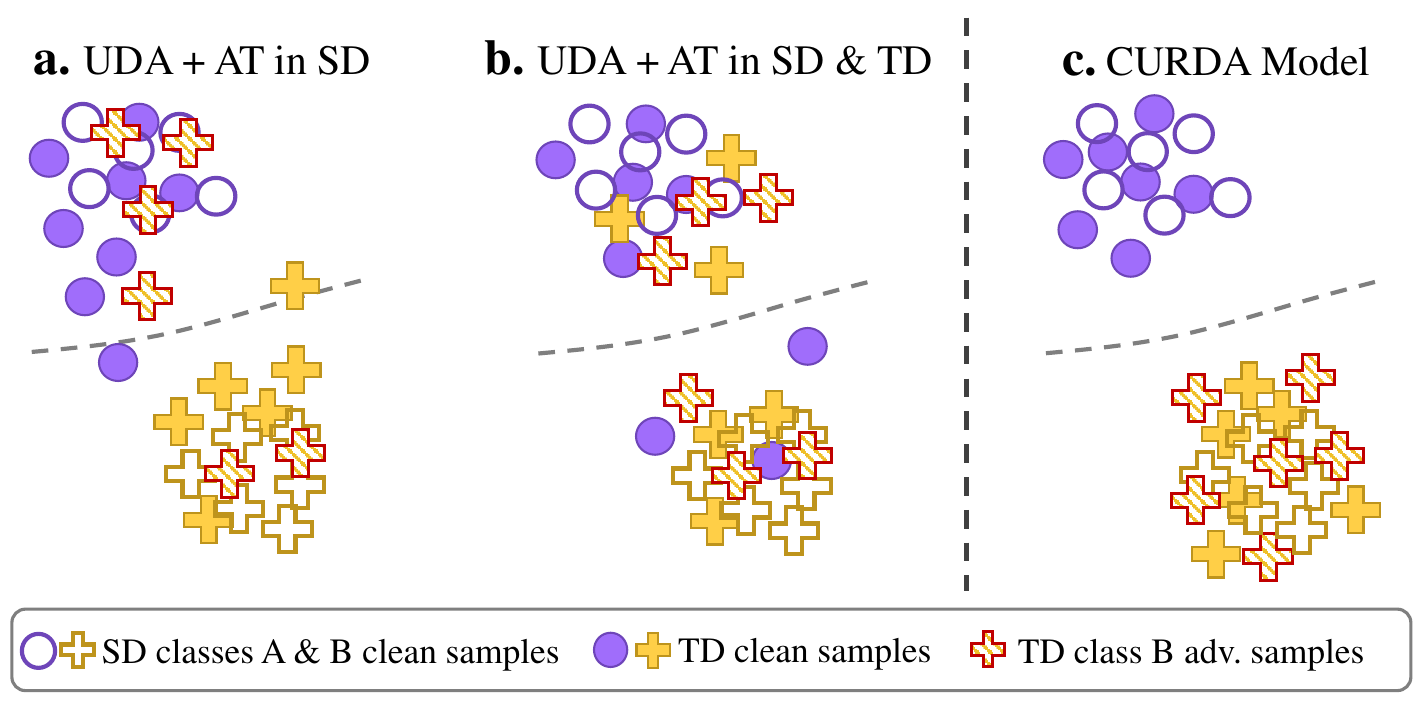}}
\caption{Illustration of performance on clean and adversarial samples under different models.
\textbf{a.} The robustness achieved by adversarial training (AT) on the source domain (SD) is hard to be transferred to the target domain(TD).
\textbf{b.} Further incorporating label-free robust training on the target domain will incur severe performance drop.
\textbf{c.} CURDA efficiently aligns both target clean and adversarial samples with their corresponding classes.
}
\label{fig:intro}
\end{figure}

Many supervised learning models suffer from significant performance drop when they are applied to a target domain with different data distribution from the labeled training source domain~\cite{wilson2020survey}.
To tackle this challenge, unsupervised domain adaptation (UDA) transfers knowledge from the labeled source domain to the unlabeled target domain by reducing the domain shift, which is commonly achieved by
aligning source and target domain data distributions
~\cite{Ganin2016DANN,kang2019contrastive,shu2018dirt,tang2020unsupervised,tzeng2017adversarial}. UDA has succeeded to generalize model performance from the source domain to the target domain in many applications~\cite{ghafoorian2017transfer,jimenez2016unsupervised,zhang2017curriculum,zou2018unsupervised}. However, it has come to our attention that the existing UDA methods overlook the network stability issues and thus the adapted deep learning model can be highly vulnerable even the model is robust in the source domain.

Adversarial robustness of deep learning models has been intensively studied in the past few years. Many works have shown that deep learning models are vulnerable to adversarial examples that contain imperceptible perturbations specifically designed to attack trained models~\cite{CW2017evaluate,Goodfellow2014advsamples,Goodfellow2017RealAdv,Madry2018Towards,Szegedy2014intriguing,Jordan2019TRADE}.
While various robust training approaches have been proposed to make deep learning models resilient to adversarial attacks~\cite{Goodfellow2018ALP,Madry2018Towards,Jordan2019TRADE},
they assume that adversarial samples in the training and testing phases are from the same distribution, \textit{i.e.}, in a same domain. When such an assumption is violated, as in the case of UDA, the robustness achieved by supervised robust training on the source domain may not be transferred together with the gained knowledge. 
Our empirical studies show that UDA methods generally align the target domain with the source domain, but a slight distribution deviation remains. Adversarial attacks can leverage such a deviation to fool the UDA model with adversarial training in source domain, as illustrated in Fig.~\ref{fig:intro}a (corresponding experiments presented in Sec.~\ref{sec:main_results}).

One natural approach to further enhance UDA models' robustness is to regularize their robustness in both the source and target domains simultaneously. Since the target domain is unlabeled, only label-free robust training methods can be employed~\cite{alayrac2019labels,carmon2019unlabeled,Goodfellow2018ALP,Jordan2019TRADE}.
Such methods have been successfully applied to supervised and semi-supervised tasks by enforcing the prediction consistency between pairs of clean-adversarial samples with metrics like KL-divergence or L2 distance. 
However, UDA is quite different from semi-supervised learning. Because the unlabeled data is in a different domain (target domain) than the labeled source domain data.
Thus, the models' performance on the unlabeled target data may not be ensured by the supervised training on the labeled source data.
Furthermore, since the adversarial samples are designed to fool the UDA model, merely enforcing the consistency between the target clean and adversarial sample pairs could mislead the domain adaptation training and cause the model to map clean samples into the wrong classes. That will result in a significant performance drop on the target clean data.

In this paper, to robustify unsupervised domain adaptation models, a new framework called Class-consistent Unsupervised Robust Domain Adaptation (CURDA) is presented.
First, we propose Contrastive Robust Training (CORTrain) in source domain to train a source-robust model, which can map input images into a more discriminative and robust space. The objective is to make the intra-class distance significantly smaller than the inter-class distance, so that the decision boundary will have a larger margin. Previous studies have shown that learning such a space can help both adversarial defense and UDA~\cite{mustafa2019adversarial,kang2019contrastive}.
Second, we introduce a Source Anchored Adversarial (SAA) contrastive loss to conduct class-aware adversarial domain alignment in UDA. 
SAA-contrastive loss is a specialized contrastive loss which utilizes source domain robust representations learned by CORTrain as fixed anchors and either pulls target domain clean and adversarial representations towards anchors or pushes them away, depending on those representations' potential labels. With SAA-contrastive loss, we not only implicitly regularize the distance between the target domain clean samples and their counterpart adversarial samples but also explicitly minimize the distribution deviation. This helps robustify UDA while preserving the performance on target clean data, as shown in Fig.~\ref{fig:intro}c.
In addition, since target domain labels are not available in UDA, we propose a strategy to generate pseudo labels in the target domain for SAA-contrastive loss. 

\begin{figure*}[t]
\centering
\includegraphics[width=1\linewidth, clip=true, trim=0 0 0 0 ]{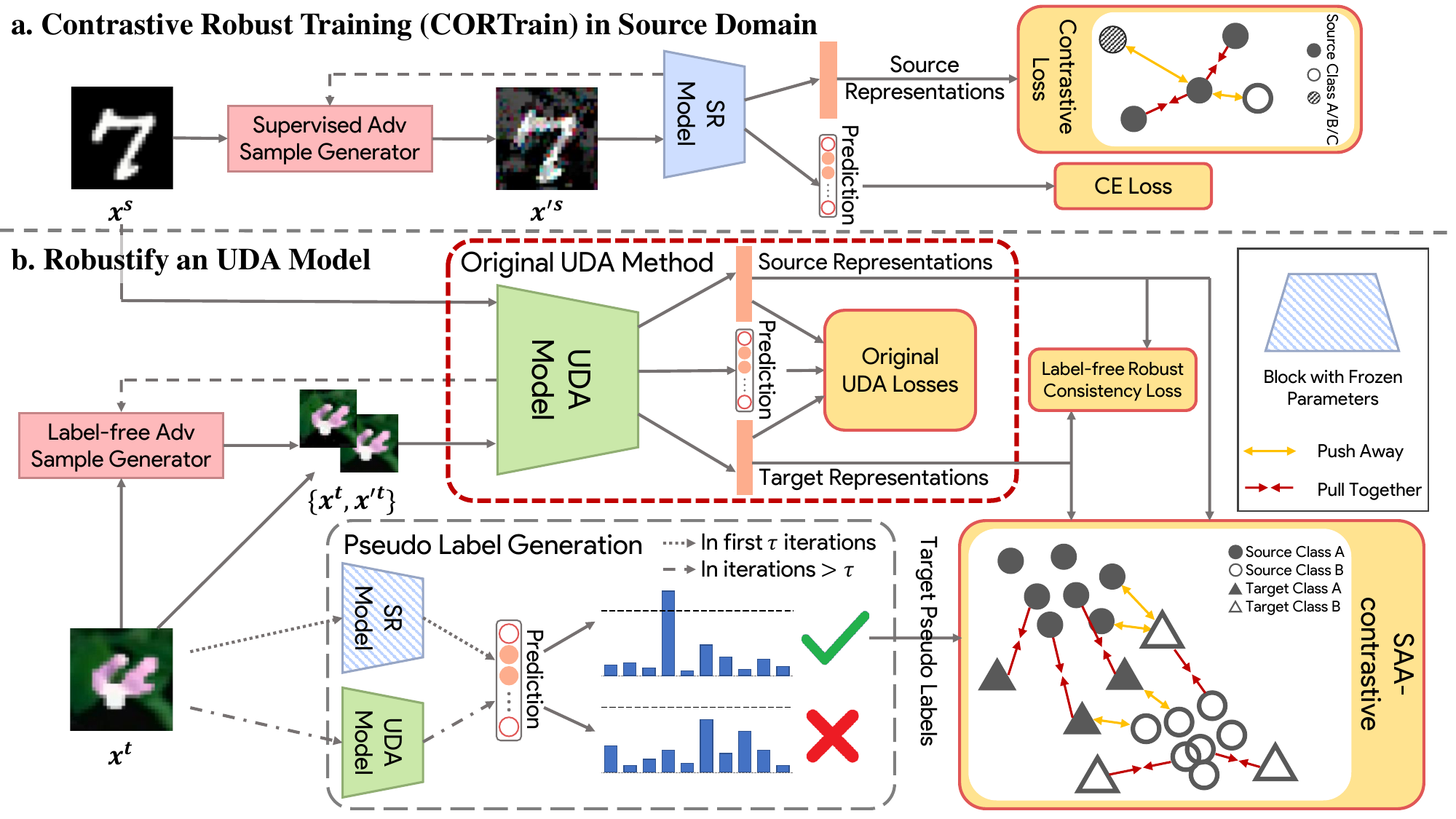}
\caption{The architecture of the proposed Class-consistent Unsupervised Robust Domain Adaptation (CURDA). The SR Model stands for a source-robust model. The UDA Model stands for a UDA model to be robustified. }
\label{fig:arch}
\end{figure*}

The experimental results show that CURDA can efficiently align both unlabeled target clean and adversarial samples to their corresponding classes, which significantly improves UDA models' robustness in the target domain while maintaining competitive classification accuracy. The main contributions of our work are summarized as follows.\\
\textbf{1)} To the best of our knowledge, this is the first work focusing on robust unsupervised domain adaptation. Our analysis shows that robustifying UDA is critical but challenging due to the inevitable domain distribution deviation in UDA.\\
\textbf{2)} We propose a new framework called \textit{Class-consistent Unsupervised Robust Domain Adaptation} (CURDA) with the \textit{Contrastive Robust Training} (CORTrain) in source domain and the \textit{Source Anchored Adverasial contrastive loss} (SAA-contrastive loss) in target domain to robustify existing UDA models.\\
\textbf{3)} The evaluations based on two different UDA models (ADDA~\cite{tzeng2017adversarial} and SRDC~\cite{tang2020unsupervised}) on two public benchmarks (DIGITS~\cite{Hull1994USPS,LeCun1998MNIST,Ganin2016DANN} and Office31~\cite{saenko2010adapting}) demonstrate that the proposed CURDA can significantly robustify UDA models yet maintain highly competitive accuracy.

\section{Related Works}

\textbf{Unsupervised domain adaptation.} 
A mainstream of UDA approaches align source and target domains by learning domain invariant feature representations~\cite{wilson2020survey}.
A typical group of these approaches leverage a domain-adversarial task to find domain–invariant representations~\cite{Ganin2015UDA,Ganin2016DANN,tzeng2017adversarial}. For example, DANN~\cite{Ganin2015UDA,Ganin2016DANN} uses adversarial training to find domain–invariant representations by a single encoder. ADDA~\cite{tzeng2017adversarial} maps source and target samples to a domain–invariant feature space with two separate encoders.
Another group of works are the discrepancy-based methods. These methods directly minimizes the domain shift measured by domain discrepancies~\cite{Tzeng2014Deepconfusion, Long2015TransFeat,Sun2016CORAL}, which may be quantified by the maximum mean discrepancy (MMD) in practice~\cite{Gretton2012Kernel}.
Some more recent works, such as  DIRT-T\cite{shu2018dirt}, CAN\cite{kang2019contrastive}, SRDC\cite{tang2020unsupervised} and CAT\cite{deng2019cluster}, show that alleviating the intrinsic inter-class mismatch benefits the model adaptation performance. For instance, different from explicit feature space regularization, SRDC~\cite{tang2020unsupervised} achieved state of the art results by assuming structural similarity between the two domains.

\textbf{Adversarial attack and defense.}
Szegedy et al.~\cite{Szegedy2014intriguing} first reported that deep neural networks can be fooled by adversarial samples, which heralded the era of adversarial attacks and model robustness for deep learning. Soon after that, Goodfellow et~al.~\cite{Goodfellow2014advsamples} proposed the fast gradient sign method (FGSM) to efficiently find such adversaries and presented a robust training approach by adding adversaries into neural network training to defence such attacks. More effective attack and defence approaches, including C\&W~\cite{CW2017evaluate}, PGD~\cite{Madry2018Towards}, BIM~\cite{Goodfellow2017RealAdv}, MIM~\cite{Dong2018Boosting}, DeepFool~\cite{Anh2015DeepFool}, and JSMA~\cite{Papernot2015JSMA}, were later proposed to identify the instabilities of deep neural networks.

Most of the robust training approaches require labels. For example, Kannan et~al.~\cite{Goodfellow2018ALP} proposed the first label-free robust training strategy, adversarial logit pairing (ALP). It applies an additional loss term constraining the pairwise logit feature distance between a clean-adversarial sample pair. Later, TRADES~\cite{Jordan2019TRADE} proposed a trade-off-inspired adversarial defense via surrogate-loss minimization, which measures model accuracy on clean samples and model robustness on adversarial samples with two separate loss terms derived theoretically. Label-free robust training approaches have been attempted for various semi-supervised learning tasks~\cite{alayrac2019labels,carmon2019unlabeled}.

\textbf{Metric learning in adversarial defense and UDA.}
Our work also relates to metric learning methods that explicitly enforce intra-class compactness and inter-class separability of the extracted features. Two of the widely used metric learning strategies are contrastive loss \cite{LeCun2006tSNE} and triplet loss \cite{Schroff2015Triplet}. Several previous works have applied these metric learning methods in adversarial attack and defense \cite{Li2019Metric,Mao2019Metric} and unsupervised domain adaptation \cite{kang2019contrastive,laiz2019using}. Our proposed SAA-contrastive loss can be considered as a novel variant of the contrastive loss to align target clean and adversarial samples with their corresponding source class samples.

\section{Method}

Fig.~\ref{fig:arch} shows the overall architecture of the proposed Class-consistent Unsupervised Robust Domain Adaptation (CURDA), which consists of two major steps. 
First, an enhanced source-robust model $F^{sr}(\cdot)$ is trained with the Contrastive Robust Training (CORTrain) in the source domain. CORTrain helps the source-robust model learn a discriminative robust space, which benefits both adversarial defense and UDA.
In the second step, the SAA-contrastive loss works together with the original UDA losses to train a robust UDA Model $F^{uda}(\cdot)$. During this step, a pseudo label generation strategy is applied to generate reliable pseudo labels to use in the SAA-contrastive loss.

Formally, let $\mathcal{S} = \{(x_1^s, y_1^s), ..., (x_{N_s}^s, y_{N_s}^s)\}$ denote a labeled source domain dataset and $\mathcal{T} = \{x_1^t, ..., x_{N_t}^t\}$ be an unlabeled target domain dataset, where $x$ represents an input data sample and $y$ represents a label. The goal of robust UDA is to train a network, which can not only perform well on clean inputs $x^t$ but also be robust against adversarial attacks $x'^t$ at test time on the target domain data $\mathcal{T}$. Here, $x'^t=x^t+\eta$ denotes an adversarial sample of $x^t$, where $||\eta||_p<\epsilon$ is an imperceptible perturbation. As this paper focuses on classification tasks, we use $y\in\{1, ..., M\}$ to denote a data label of $M$ classes. For simplicity, we use $F_c^{sr}(x^s)$ and  $F_c^{uda}(x^t)$ to represent the predicted classifications of the models. We then use $F_f^{sr}(x^s)$ and $F_f^{uda}(x^t)$ to denote the feature representations learned by the models.

\subsection{Contrastive Robust Training}
\label{sec:cortrain}

Mapping data into a discriminative representation space with intra-class compactness and inter-class separability has been proven to benefit both adversarial defense and UDA~\cite{Li2019Metric,Mao2019Metric,kang2019contrastive}.
Thus, in order to robustify a UDA model, we first propose Contrastive Robust Training (CORTrain) to learn such a discriminative representation space in the source domain.

Specifically, an adversarial sample $x'^s$ of a labeled source sample $x^s$ can be generated by a supervised adversarial sample generator, which is the projected gradient descent (PGD)~\cite{Madry2018Towards} attack in our work. Then two losses, the cross entropy loss $\mathcal{L}_{ce}$ and the contrastive loss $\mathcal{L}_{con}$, are minimized on the input samples $x'^s$ and corresponding labels $y^s$ to train the classifier while constraining the intra-class compactness and inter-class sparsity. The cross entropy loss, $\mathcal{L}_{ce}=-\sum_{k=1}^M\mathbbm{1}_{[k=y^s]}\log F^{sr}_c(x'^s)$, guides $F^{sr}(\cdot)$ to map the adversarial sample onto the right label. The contrastive loss imposes constraint on the learned representations $F^{sr}_f(x'^s)$ to minimize the intra-class distance and maximize the inter-class distance. For a pair of samples in a mini-batch, $\{(x_i'^s, y_i^s), (x_j'^s, y_j^s)\}$, the contrastive loss is defined as
\begin{equation}
\label{eq:cortrain}
    \mathcal{L}_{con} = \frac{1}{2} [\mathbbm{1}_{[y_i^s=y_j^s]}D_r^2
    +\mathbbm{1}_{[y_i^s\ne y_j^s]}{\max(0, m_s-D_r)}^2],
\end{equation}
where $D_r=||F^{sr}_f({x'_i}^s) - F^{sr}_f({x'_j}^s)||_2$ is the Euclidean distance between two feature representations, and $m_s>0$ is a margin to prevent over-fitting. The complete loss function for CORTrain is given by $\mathcal{L}_{cort}=\mathcal{L}_{ce} + \lambda_{con}\mathcal{L}_{con}$, where $\lambda_{con}$ is a positive weighting parameter.

\subsection{Source Anchored Adversarial Contrastive Loss}
\label{sec:saa}

To robustify a UDA model, the Source Anchored Adversarial (SAA) contrastive loss is proposed to work together with the original UDA losses in the domain adaptation. As shown in Fig.~\ref{fig:arch}b, the SAA-contrastive loss uses source data representations as anchors, and optimizes the inter- and intra-class distances between target domain representations (of both clean and adversarial samples) and source domain representations. It is worth noting that the source-robust model $F^{sr}(\cdot)$ keeps frozen during the process. On one hand, by pulling the target representations of both clean and adversarial samples towards the source anchors of their own classes, the SAA-contrastive loss effectively minimizes the distance between target clean-adversarial pairs in an implicit way. This is different from directly minimizing the distance between the target clean-adversarial pairs.
Because the source anchors are fixed, the SAA-contrastive loss prevents adversarial samples from dragging clean samples towards wrong classes, which would otherwise be the case under direct minimization of the distance. On the other hand, by pushing the target representations away from the source anchors of different classes, the SAA-contrastive loss enlarges the margin of the decision boundary, which makes the model harder to attack.

For a pair of data $\{(u, y_u), (v, y_v)\}$, one term of the SAA-constrastive loss is formulated similar as a standard contrastive loss:
\begin{align}
\label{eq:saa_1}
    l(u,&v,y_u,y_v) = \nonumber\\
    &\frac{1}{2} \times
    \begin{cases}
        D(u,v)^2, & \text{if $y_u=y_v$} \\
        {max(0, m_t-D(u, v))}^2, & \text{if $y_u\ne y_v$}
    \end{cases} 
\end{align}
where $D(\cdot,\cdot)$ denotes the Euclidean distance and $m_t>0$ is a margin to prevent over-fitting. 
Since target labels are not available during UDA, to compute the SAA-contrastive loss, we introduce a strategy to generate pseudo labels $\hat{y}^t$ for target samples. In the first $\tau$ iterations of adaptation training, the performance of $F^{uda}(\cdot)$ may be unstable, so the source-robust model $F^{sr}(\cdot)$ with frozen parameters is used for pseudo label generation. To reduce the noise in the pseudo labels, only predictions with the largest softmax scores greater than a specific threshold $P_{pseudo}$ are preserved. Samples with their largest softmax scores less than $P_{pseudo}$, indicating uncertain estimation, would not be included in the SAA-contrastive loss. After the first $\tau$ iterations, when the UDA modelbecomes stable,  $F^{uda}(\cdot)$ is switched to generate pseudo labels.

The full SAA-contrastive loss is formulated as
\begin{align}
\label{eq:saa}
    \mathcal{L}_{saa} =
    &\mathbb{E}_{\{(x^s,y^s), x^t\}\sim\{\mathcal{S}, \mathcal{T}\}}[l(F^{sr}_f(x^s), F^{uda}_f(x^t), y^s, \hat{y}^t) \nonumber\\
    &~~~~~~~~~~~
    +l(F^{sr}_f(x^s), F^{uda}_f(x'^t), y^s, \hat{y}^t)],
\end{align}
where $x'^t$ denotes an adversarial sample of a target clean sample $x^t$ generated by a label-free adversarial sample generator based on $F^{uda}(\cdot)$.
In our paper, we used the adversarial sample generator proposed in~\cite{Jordan2019TRADE}. 

\subsection{Other Losses}

\textbf{Original UDA losses.} In this paper, we evaluated robustness enhancement on two different UDA models, \textit{i.e.}, ADDA~\cite{tzeng2017adversarial} and SRDC~\cite{tang2020unsupervised}. 
The former was trained to adapt the source domain model $F^{sr}(\cdot)$ to the target domain model $F^{uda}(\cdot)$ using the adversarial-discriminative loss.
The latter uses exactly the same network described in the original paper~\cite{tang2020unsupervised} and also the Structurally Regularized Deep Clustering (SRDC) loss to train the UDA model $F^{uda}(\cdot)$.
Since these orignal UDA losses are not our contributions in this work, we simply use $\mathcal{L}_{uda}$ to denote them. Their detailed formulas are provided in the supplementary material.

\textbf{Label-free robust consistency loss.} 
Since the proposed SAA-contrastive loss uses only those target samples with reliable pseudo labels, to make full use of all the unlabeled samples in the target domain, we further include the label-free robust consistency losses $\mathcal{L}_{trade}$ presented in TRADES~\cite{Jordan2019TRADE} to regularize the consistency between target clean and adversarial samples.
It is given as
\begin{equation}
\label{eq:trades}
\mathcal{L}_{trade} = -\mathbb{E}_{x^t\sim\mathcal{T}}KL(C(E_{uda}(x^t), C(E_{uda}(x'^t))
\end{equation}
where $KL(\cdot, \cdot)$ denotes the Kullback-Leibler divergence. 
As mentioned in Sec.~\ref{sec:saa}, the SAA-contrastive loss can prevent the adversarial samples from misleaing the domain adaptation training. Therefore, this label-free robust consistency loss can instead further enhance the model robustness by exploiting more target domain data.

The overall adaptation loss of CURDA is
\begin{equation}
\label{eq:total}
\mathcal{L}_{curda} = \lambda_{saa}\mathcal{L}_{saa} + \mathcal{L}_{uda} + \mathcal{L}_{trade},
\end{equation}
where $\lambda_{saa}$ is a positive weighting parameter.

\section{Experiments}

The proposed CURDA framework were evaluated across two public benchmarks, i.e., DIGITS~\cite{Hull1994USPS,LeCun1998MNIST,Ganin2016DANN} and Office-31~\cite{saenko2010adapting}. UDA models were evaluated by both clean data classification \emph{accuracy} and adversarial \emph{robustness} in the target domain. 
In the standard experiments for model robustness evaluation~\cite{Madry2018Towards,Jordan2019TRADE}, the adversarial samples generated from the training set cannot be used in the test set. Although under the scenario of UDA, the model did not have access to the target domain labels, it learned the prior of the clean-adversarial sample pairs through the label-free robust training consistency loss and the SAA-contrastive loss.
Thus, we designed several experiments following the settings in the inductive UDA\cite{wilson2020survey,tang2020unsupervised}. We followed a 50\%/50\% split scheme to divide each target domain dataset into the training and test sets. We used the labeled dataset of the source domain and the unlabeled training set of the target domain as the training data. The reported test results were obtained by using the best performing model selected with the target training set.

The rest of this section first presents the experimental results and qualitative visualizations on the DIGITS and Office-31 benchmarks to compare CURDA with three baseline models, i.e., the origin UDA model, UDA with adversarial training in the source domain(UDA+AT in SD) and UDA with adversarial training in the source and target domains (UDA+AT in SD \& TD).
Then, ablation studies were performed on Office-31 benchmark to evaluate the effectiveness of each component in CURDA. In all of our quantitative analyses, the \emph{robustness} was evaluated by the classification accuracy on the adversarial samples generated with PGD attack\cite{Madry2015AT}. 
The same set of experiments with the same settings were also conducted on the other two popular datasets, \textit{i.e.}, VisDA-2017~\cite{peng2017visda} and CIFAR-STL~\cite{Krizhevsky2009Cifar,coates2011analysis}. Considering that the experimental results on these two datasets are similar with those on the DIGITS and Office-31 datasets, we put them in the supplementary material.

\subsection{Datasets}

\textbf{Handwritten digits (DIGITS)} includes three data domains, i.e., MNIST~\cite{LeCun1998MNIST}, USPS~\cite{Hull1994USPS} and MNIST-M~\cite{Ganin2016DANN}. The MNIST dataset is also frequently used for the evaluation of adversarial attack and defence. 
The number of samples are imbalanced across the three domains, with 60,000 binary images in MNIST, 7,291 binary images in USPS and 68,002 RGB images in MNIST-M. 
On DIGITS, we conducted three domain adaptation tasks, including MNIST$\rightarrow$USPS, USPS$\rightarrow$MNIST and MNIST$\rightarrow$MNIST-M.

\textbf{Office-31}~\cite{saenko2010adapting} is composed by 4,110 images from three different domains, i.e., \textbf{A}mazon (\textbf{A}), \textbf{W}ebcam (\textbf{W}) and \textbf{D}SLR (\textbf{D}). The dataset is imbalanced across domains, with 2,817 images in domain A, 795 images in domain W, and 498 images in domain D. There are a total of 31 classes shared across domains. Experiments were conducted on all 6 source-target domain pairs constructed by these three domains.

\subsection{Baseline Models}
\label{sec:baseline}

Since to the best of our knowledge, this is the first work focuses on robustifying UDA models, we present three baseline models to compare with the proposed CURDA.

\textbf{UDA model only.} This baseline model does not use any adversarial training strategies. It validates the hypothesis that regular UDA models are vulnerable to target domain adversarial samples.

\textbf{UDA with adversarial training in the source domain (UDA+AT in SD).} The adversarial training strategy of PGD~\cite{Madry2018Towards} was used to robustify the existing UDA models through source domain adversarial training. This baseline model validates the hypothesis that the source domain robustness may not directly be transferred to the target domain.

\textbf{UDA with adversarial training in the source and target domains (UDA+AT in SD \&TD).} The third baseline model further introduces the label-free robust training on the target domain. It intends to evaluate whether label-free robust training could enhance the target domain robustness.

\subsection{Implementation Details}

\textbf{Backbone architectures.}
Following the same network architecture as in~\cite{Madry2015AT,tang2020unsupervised,tzeng2017adversarial}, we used a modified LeNet~\cite{LeCun1998MNIST} and ResNet-50\cite{he2016deep} as the backbone encoder in DIGITS and Office-31 benchmarks, respectively. The modified LeNet composes of two convolutional layers followed by two fully-connected layer. For the domain discriminator in ADDA, we used the same structure described in the original paper~\cite{tzeng2017adversarial}.

\textbf{Training details.}
Stochastic gradient descent (SGD) is used as the optimizer in all training processes with a momentum of $0.9$. For each experiment, we trained the model from scratch three times with different random seeds and report averaged performance in the form of $mean(\pm std)$. On DIGITS, in the first step of CURDA, the source-robust model was trained for $80$ epochs. The learning rate is $0.001$ for the first $50$ epochs and decreases to $0.0001$ for the rest of $30$ epochs. For the adaptation step, the UDA encoder is trained with a learning rate of $0.001$ for $60$ epochs. A weight decay of $5e-4$ and a batch size of $300$ is used in both two steps. For pseudo label generation, we used $\tau=20$ and $P_{pseudo}=0.90$. In the losses of $\mathcal{L}_{con}$ and $\mathcal{L}_{saa}$, the margins $m_s$ and $m_t$ were set to be 10. 
We set the weighting parameters $\lambda_{con}=0.01$ and $\lambda_{saa}=0.01$ in the loss functions of CORTrain (Eq.\ref{eq:cortrain}) and SAA (Eq.\ref{eq:saa}), respectively.

On Office-31, the source-robust model was trained for $40$ epochs with a learning rate of $0.001$ and $0.0001$ for the first $30$ and the remaining $10$ epochs, respectively. The UDA encoder was trained for $40$ epochs with a learning rate of $0.001$. A weight decay of $2e-4$ and a batch size of $128$ were used for both source model training and adaptation. The pseudo label generator used $\tau=30$ and $P_{pseudo}=0.85$. The margins $m_s$ and $m_t$ were both set to be 25. In the loss functions, we set $\lambda_{con}=\lambda_{saa}=0.01$.

\textbf{Adversarial attacks.} 
The PGD~\cite{Madry2018Towards} attack was used for both CORTrain in the source domains and robustness testing in the target domains. The hyper-parameters for training and testing are consistent for each domain following the popular settings used in the existing works~\cite{Jordan2019TRADE,wong2018scaling}. For binary image datasets, MNIST and USPS, the attack perturbation is $0.3$, the perturbation step size is $0.01$ and the number of perturbation steps is $40$. For colorful image datasets, MNIST-M and Office-31, the attack perturbation is $0.031$, the perturbation step size is $0.007$ and the number of perturbation steps is $40$. For the label-free adversarial samples generated by TRADES~\cite{Jordan2019TRADE}, it shares the same hyper-parameter setups with PGD on the source domain.

\subsection{Main Results}

\label{sec:main_results}
\begin{table*}[t!]
	\centering
	\caption{Clean data accuracy (Clean (\%)) and adversarial data robustness (Rob. (\%)) on the DIGITS benchmark. The best results on each task are shown in \textbf{bold} in each column.}
	\label{tab:Digits}
	\scalebox{0.80}{
	\begin{tabular}{|l|c||c|c||c|c||c|c|}
		\hline
		\multirow{2}{*}{\textbf{UDA models}} & \multirow{2}{*}{\textbf{AT methods}} & \multicolumn{2}{c||}{\textbf{MNIST$\rightarrow$USPS}} &  \multicolumn{2}{c||}{\textbf{USPS$\rightarrow$MNIST}} & \multicolumn{2}{c|}{\textbf{MNIST$\rightarrow$MNIST-m}}\\
		\cline{3-8}
		& & Clean & Rob. & Clean & Rob. & Clean & Rob.\\
		\hline
		\hline
		\multirow{3}{*}{\textbf{ADDA}} & None & {\bf89.4}$\pm$0.1 & 0.0$\pm$0.0 & {\bf90.0}$\pm$0.1 & 0.0$\pm$0.0 & {\bf78.0}$\pm$0.1 & 0.0$\pm$0.0 \\ 
		\cline{2-8}
		& AT in SD & 86.9$\pm$0.2 & 10.1$\pm$0.1 & 83.7$\pm$0.2 & 12.3$\pm$0.3 & 71.3$\pm$0.2 & 9.3$\pm$0.4 \\ 
		\cline{2-8}
		& AT in SD \& TD & 82.4$\pm$0.2 & 63.4$\pm$0.3 & 75.4$\pm$0.2 & 54.2$\pm$0.3 & 56.7$\pm$0.3 & 36.4$\pm$0.3 \\
		\cline{2-8}
		& CURDA & 86.8$\pm$0.2 & {\bf77.5}$\pm$0.2 & 85.3$\pm$0.2 & {\bf76.6}$\pm$0.2 & 66.8$\pm$0.2 & {\bf60.6}$\pm$0.3 \\ 
		\hline
		\hline
		\multirow{3}{*}{\textbf{SRDC}} & None & {\bf94.8}$\pm$0.2 & 0.0$\pm$0.0 & {\bf96.0}$\pm$0.1 & 0.0$\pm$0.0 & {\bf86.3}$\pm$0.2 & 0.0$\pm$0.0 \\
		\cline{2-8}
		& AT in SD & 89.6$\pm$0.2 & 65.2$\pm$0.3 & 86.1$\pm$0.1 & 76.2$\pm$0.3 & 78.8$\pm$0.2 & 34.5$\pm$0.3 \\
		\cline{2-8}
		& AT in SD \& TD & 86.1$\pm$0.1 & 76.2$\pm$0.2 & 82.2$\pm$0.1 & 70.8$\pm$0.2  & 68.5$\pm$0.2 & 54.2$\pm$0.1  \\ 
		\cline{2-8}
		& CURDA & 93.6$\pm$0.1 & {\bf90.1}$\pm$0.1 & 94.9$\pm$0.1 & {\bf92.2}$\pm$0.1 & 83.2$\pm$0.2 & {\bf76.8}$\pm$0.2 \\ 
		\hline
	\end{tabular}
	}
\end{table*}

\textbf{DIGITS}
Tab.~\ref{tab:Digits} presents the comparison of the proposed CURDA with three baselines on the DIGITS benchmark. 
It can be seen that without adopting any adversarial training strategies, \textit{ADDA and SRDC} cannot defend against target domain adversarial samples. Compared with UDA models, CURDA significantly improved the robustness by more than $60\%$. The slight degradation of clean sample accuracy is unsurprising since multiple work has discussed the trade-off between accuracy and robustness~\cite{Jordan2019TRADE, Rozsa2016AccRob}. Since we performed experiments in inductive UDA scheme, the UDA performance is slightly lower than the original paper.

It is worth noting that incorporating UDA with adversarial training in source domain, \textit{i.e., UDA+AT in SD}, improved the performance to certain extent. However, our proposed CURDA outperformed it on robustness and maintained a competitive accuracy on clean data in the same time.
In addition, compared with \textit{UDA+AT on SD}, \textit{UDA+AT on SD \& TD} achieves better robustness in most of the cases, but suffers from accuracy drop on clean samples. This result verifies that the label-free robust loss can mislead the domain adaptation training and cause misclassification on target clean samples.

\begin{figure}
\centering
{\includegraphics[width=.48\textwidth, clip=true, trim=0 0 0 0]{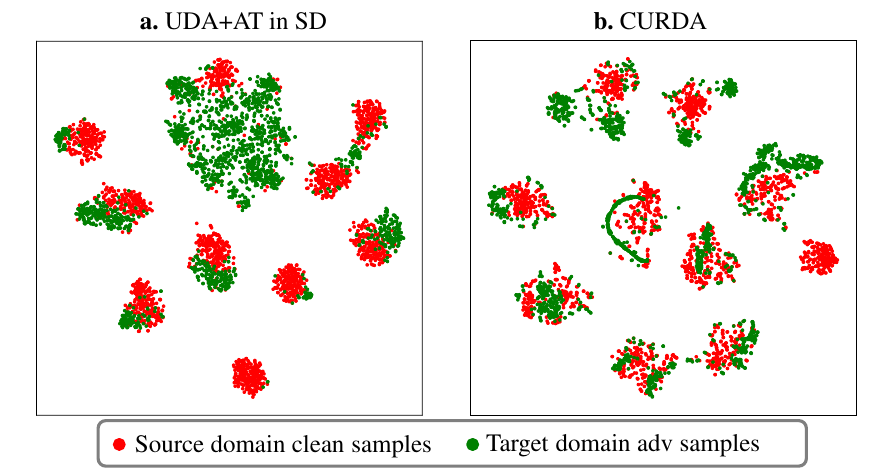}}
\hfill
\caption{Visualization by t-SNE for \textbf{a.} \textit{UDA+AT on SD} and \textbf{b.} CURDA. The outputs of the second-to-last layer (activation layer) were used for the visualization. The source and target domains here are MNIST (red) and USPS (green), respectively.}
\label{fig:demo_1}
\end{figure}

\begin{figure*}[t]
\centering
\includegraphics[width=\textwidth, clip=true, trim=0 0 0 0]{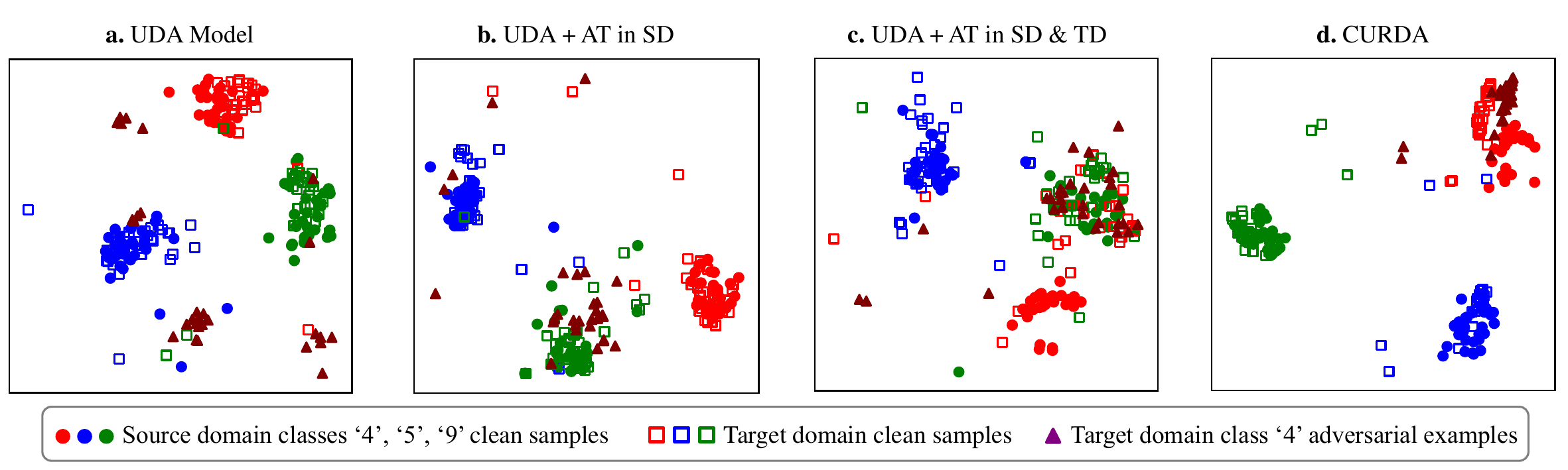}
\caption{Visualization by t-SNE on representations learnt by ADDA with different adversarial training strategies. The outputs of the second-to-last layer (activation layer) were used for this visualization.
}
\label{fig:demo_2}
\end{figure*}

To further investigate the difference between CURDA and baseline models, we visualized the learnt representations with t-SNE~\cite{LeCun2006tSNE} on MNIST $\rightarrow$ USPS task in Fig.~\ref{fig:demo_1} and ~\ref{fig:demo_2}. Fig.~\ref{fig:demo_1} shows the feature space distribution of two ADDA models trained with \textit{UDA+AT in SD} and CURDA, respectively. In \textit{UDA+AT in SD} most of the target adversarial samples are not matched with source samples. In contrast, the target adversarial samples are close to the corresponding source sample clusters in CURDA's representation space, which is mainly due to the combination of the CORTrain and the SAA-contrastive loss.

Fig.~\ref{fig:demo_2} visualize the feature presentations learnt by ADDA with different adversarial training strategies. We visualized the representation of three digits classes (`4', `5' and `9').
In Fig.~\ref{fig:demo_2}a, it can be observed that without any adversarial training, the representation of the target adversarial data `4' is clustered wrongly with other classes. It is noteworthy that on the figure it seems like some of these adversarial data clustered on some empty location, but these locations are actually occupied by some classes other than `4', `5' and `9'.  It verifies that without adversarial training UDA models present no defense against target domain adversarial attacks.
Fig.~\ref{fig:demo_2}b. shows the visualization of \textit{UDA+AT in SD}. Although adversarial training was applied on the source domain, we did not observe much improvement of target domain robustness (most of target adversarial data '4' are misclassified). This indicates that the robustness achieved by supervised robust training on the source domain is hard to be transferred to the target domain. Adversarial attacks can leverage any slight domain distribution deviation to fool UDA models.
Fig.~\ref{fig:demo_2}c further demonstrates that incorporating label-free robust training with UDA will mislead the domain adaptation training. As we can see that although the target adversarial samples (solid triangles) get closer to their corresponding clean samples (red hollow squares), both of them are mixed with green hollow squares (the target domain `9'). 
However, in Fig~\ref{fig:demo_2}d, we can see that the proposed CURDA successfully aligned the target clean and adversarial samples with those source samples of the same classes, which justifies both high accuracy and strong robustness achieved by CURDA.

\begin{table*}[t!]
	\centering
	\caption{Clean data accuracy (Clean (\%)) and adversarial data robustness (Rob. (\%)) on Office-31 benchmark. The bold number shows the best results.}
    \label{tab:Office31}
	\scalebox{0.66}{
	\begin{tabular}{|l|c||c|c||c|c||c|c||c|c||c|c||c|c|}
		\hline
		\multirow{2}{*}{\textbf{UDA models}} & \multirow{2}{*}{\textbf{AT methods}} & \multicolumn{2}{c||}{\textbf{A$\rightarrow$W}} &  \multicolumn{2}{c||}{\textbf{W$\rightarrow$D}} & \multicolumn{2}{c||}{\textbf{D$\rightarrow$A}} & \multicolumn{2}{c||}{\textbf{A$\rightarrow$D}} & \multicolumn{2}{c||}{\textbf{D$\rightarrow$W}} & \multicolumn{2}{c|}{\textbf{W$\rightarrow$A}}\\
		\cline{3-14}
		& & Clean & Rob. & Clean & Rob. & Clean & Rob. & Clean & Rob. & Clean & Rob. & Clean & Rob.\\
		\hline
		\hline
		\multirow{3}{*}{\textbf{ADDA}} & None & {\bf90.6}$\pm$0.1 & 0.2$\pm$0.0 & {\bf99.4}$\pm$0.2 & 0.1$\pm$0.0 & {\bf61.7}$\pm$0.4 & 0.1$\pm$0.0 & {\bf85.7}$\pm$0.2 & 0.1$\pm$0.0 & {\bf95.6}$\pm$0.1 & 0.1$\pm$0.0 & {\bf60.8}$\pm$0.4 & 0.1$\pm$0.0\\ 
		\cline{2-14}
		& AT in SD & 89.4$\pm$0.3 & 18.8$\pm$0.4 & 94.5$\pm$0.3 & 35.7$\pm$0.4 & 56.7$\pm$0.4 & 9.0$\pm$0.3 & 80.1$\pm$0.2 & 16.5$\pm$0.4 & 91.1$\pm$0.2 & 31.5$\pm$0.3 & 57.8$\pm$0.2 & 7.2$\pm$0.2 \\ 
		\cline{2-14}
		& AT in SD \& TD & 46.5$\pm$0.3 & 37.7$\pm$0.4 & 97.2$\pm$0.2 & 85.5$\pm$0.2 & 29.5$\pm$0.4 & 22.1$\pm$0.3 & 47.0$\pm$0.3 & 33.5$\pm$0.2 & 90.6$\pm$0.2 & 81.5$\pm$0.2 & 31.6$\pm$0.3 & 20.7$\pm$0.2\\
		\cline{2-14}
		& CURDA & 88.0$\pm$0.2 & {\bf86.4}$\pm$0.2 & 97.8$\pm$0.1 & {\bf95.8}$\pm$0.2 & 61.2$\pm$0.2 & {\bf54.1}$\pm$0.2 & {\bf85.7}$\pm$0.2 & 83.5$\pm$0.1 & 94.2$\pm$0.1 & {\bf92.5}$\pm$0.2 & 60.1$\pm$0.1 & {\bf54.9}$\pm$0.2\\ 
		\hline
		\hline
		\multirow{3}{*}{\textbf{SRDC}} & None & {\bf91.7}$\pm$0.1 & 0.0$\pm$0.0 & {\bf99.7}$\pm$0.2 & 0.0$\pm$0.0 & {\bf75.6}$\pm$0.2 & 0.0$\pm$0.0 & {\bf91.6}$\pm$0.2 & 0.0$\pm$0.0 & {\bf99.2}$\pm$0.1 &  0.0$\pm$0.0 & {\bf75.7}$\pm$0.1 & 0.0$\pm$0.0 \\
		\cline{2-14}
		& AT in SD & 90.6$\pm$0.1 & 61.2$\pm$0.2 & 98.2$\pm$0.1 & 68.6$\pm$0.2 & 68.8$\pm$0.2 & 44.5$\pm$0.2 & 91.0$\pm$0.2 & 63.5$\pm$0.2 & 98.1$\pm$0.1 & 63.2$\pm$0.2 & 63.4$\pm$0.2 & 42.7$\pm$0.2 \\
		\cline{2-14}
		& AT in SD \& TD & 66.8$\pm$0.2 & 43.2$\pm$0.2 & 96.2$\pm$0.2 & 87.8$\pm$0.1 & 38.6$\pm$0.2 & 26.3$\pm$0.3 & 67.2$\pm$0.2 & 44.1$\pm$0.2 & 93.6$\pm$0.2 & 86.5$\pm$0.2 & 37.9$\pm$0.2 & 25.3$\pm$0.1\\ 
		\cline{2-14}
		& CURDA & 90.5$\pm$0.2 & {\bf89.1}$\pm$0.3 & 98.4$\pm$0.2 & {\bf96.1}$\pm$0.1 & 72.3$\pm$0.1 & {\bf66.8}$\pm$0.1 & 90.9$\pm$0.2 & {\bf88.6}$\pm$0.1 & 97.7$\pm$0.1 & {\bf95.4}$\pm$0.2 & 72.8$\pm$0.2 & {\bf67.0}$\pm$0.2\\ 
		\hline
	\end{tabular}
	}
\end{table*}

\textbf{Office-31} 
The results on the Office-31 dataset are presented in Tab.~\ref{tab:Office31}. Although SRDC has achieved the state-of-the-art performance, slight distribution deviation is hardly to avoid. \textit{UDA+AT in SD} shows that such slight inter-domain deviation would be leveraged by adversarial attacks to fool the model, such as in \textbf{W$\rightarrow$D} and \textbf{D$\rightarrow$W}. Similar to the DIGITS dataset, \textit{UDA+AT in SD \& TD} suffers from severe accuracy drop on target clean samples, which is mainly due to the weakness of label-free robust training as discussed in the experiments on DIGITS. Even worse, \textit{SRDC+AT in SD \& TD} even experienced performance drop on both clean data accuracy and robustness in \textbf{W$\rightarrow$A} and \textbf{D$\rightarrow$A}. In comparison, only the CURDA obtained steady high improvement on robustness while maintaining a competitive accuracy on clean data.

\subsection{Ablation Studies and Analysis}
Tab.~\ref{tab:Ablation} shows the results of ablation study that examining all the five componens of CURDA. We performed leave-one-component-out of our framework. Two tasks of Office-31(A$\rightarrow$W and W$\rightarrow$A) were used as examples to show the effectiveness of each component in CURDA robustified ADDA. The results of similar ablation study on CURDA robustified SRDC are shown in the supplementary material.

\textbf{Effect of CORTrain.}
Removing the $\mathcal{L}_{con}$ in CORTrain step caused a drop of accuracy on both target clean data and target adversarial data. This result verified that mapping data into a more discriminative and robust space could benefit \textit{both} adversarial defence and UDA.

\textbf{Effect of SAA-contrastive loss.}
Without the $\mathcal{L}_{saa}$, both the clean data accuracy and the robustness decreased significantly. This phenomenon demonstrates that the SAA-contrastive loss plays an important role to increase the target model robustness while keeping the accuracy of UDA models.

\textbf{Effect of label-free robust training.}
Comparing the row 'w/o. $\mathcal{L}_{trade}$' and 'CURDA' in Tab.~\ref{tab:Ablation}, we observed a slight drop by removing the loss of $\mathcal{L}_{trade}$. This suggests that our proposed SAA-contrastive loss has already forced most of the adversarial samples to be close with the counterpart clean samples, so further explicitly constraining the distance between clean-adversarial pairs didn't bring too much performance improvement.

\textbf{Effects of switching encoders in pseudo label generation.} The quality of pseudo labels determines the effectiveness of the SAA-contrastive loss. From the column `w s-lab gen' in Tab.~\ref{tab:Ablation}, it can be seen that switching encoders can bring the improvement on both clean sample accuracy and model robustness. This results justify the necessity of
switching the pseudo label generator from the pretrained source-robust model to the UDA model, when the performance of the latter is stable. Because a well-trained UDA model can generate pseudo labels with higher quality than a source domain model.

\begin{table}[ht]
\centering
\caption{Ablation study. Rows 2-4 present the effects of four major losses used in CURDA evaluated on A$\rightarrow$W and W$\rightarrow$A. 
The row '\textit{w/o. switch lab-gen}' means pseudo label generation without switching encoders.}
\scalebox{0.87}{
\begin{tabular}{|l||c|c|c|c|}
\hline
\multirow{2}{*}{Methods} & \multicolumn{2}{c|}{A$\rightarrow$W} & 
\multicolumn{2}{c|}{W$\rightarrow$A}\\
\cline{2-5}
& Clean & Rob. & Clean & Rob. \\
\hline
w/o. $\mathcal{L}_{con}$ & 84.6$\pm$0.2 & 77.2$\pm$0.2 & 51.9$\pm$0.3 & 42.3$\pm$0.3\\
\hline
w/o. $\mathcal{L}_{saa}$ & 46.5$\pm$0.3 & 37.7$\pm$0.4 & 31.6$\pm$0.4 & 20.6$\pm$0.3\\
\hline
w/o. $\mathcal{L}_{trade}$ & 86.9$\pm$0.3 & 85.4$\pm$0.2 & 59.5$\pm$0.2 & 53.6$\pm$0.2\\
\hline
w/o. switch & \multirow{2}{*}{81.4$\pm$0.2} & \multirow{2}{*}{74.4$\pm$0.3} & 
\multirow{2}{*}{56.5$\pm$0.3} & \multirow{2}{*}{49.3$\pm$0.2}\\
label gen. &  &  &  & \\
\hline
CURDA & {\bf88.0}$\pm$0.2 & {\bf86.4}$\pm$0.3 & {\bf60.1}$\pm$0.2 & {\bf54.9}$\pm$0.2\\
\hline
\end{tabular}
}
\label{tab:Ablation}
\end{table}

\subsection{Hyperparameter Sensitivity}
\begin{figure}
\centering
{\includegraphics[width=.98\linewidth]{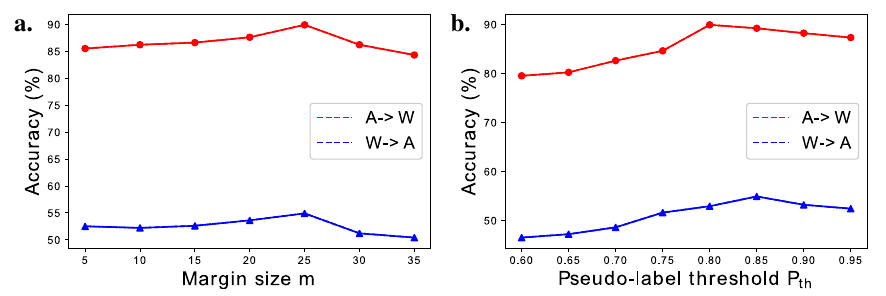}}
\caption{The hyperparameter sensitivity w.r.t. \textbf{a.} margin size $m$ ($m=m_{s}=m_{t}$) in Eq.~(\ref{eq:cortrain}) and ~(\ref{eq:saa}), \textbf{b.} pseudo label generation threshold $P_{pseudo}$. A$\rightarrow$W and W$\rightarrow$A in Office-31 are taken as examples.
}\label{fig:para_sens}
\end{figure}

We examined the hyperparameter sensitivity of our CURDA framework to margin size $m$ ($m=m_{s}=m_{t}$ in Eq.~(\ref{eq:cortrain}) and ~(\ref{eq:saa})) and pseudo label generation threshold $P_{pseudo}$. For each hyperparameter, we performed the experiments by fixing all the others to the default values. The results of A$\rightarrow$W(red) and W$\rightarrow$A(blue) on CURDA robustified ADDA are provided as examples. The trends on other tasks are similar. As shown in Fig.~\ref{fig:para_sens}a, the accuracy is relatively stable when the margin size $m$ is between 5 to 25, but dropped slightly if the margin size is larger than 25. As shown in Fig.~\ref{fig:para_sens}b, the selection of $P_{pseudo}$ will greatly affect the results. The main reason is that the threshold $P_{pseudo}$ determines the amount of correct target pseudo labels that can be used for training. We find that the model works well when $P_{pseudo}$ falls in $[0.80,0.85]$.

\section{Conclusion}
In this paper, we presented a new problem on robustifying UDA, which concerns training an adversarial robust model in a target domain without data annotations. Correspondingly, we proposed a novel framework of Class-consistent Unsupervised Robust Domain Adaptation(CURDA)
to tackle this problem. With contrastive robust training (CORTrain) to enhance the source domain training and source anchored adversarial contrastive (SAA-contrastive) loss to perform class-aware adversarial domain alignment, CURDA conquered the challenge of the inevitable domain distribution deviation in UDA against target domain adversarial training.
Experiments on DIGITS and Office-31 benchmarks demonstrate that the proposed CURDA can effectively robustify existing UDA models while simultaneously achieving strong robustness and competitive accuracy. The effectiveness of each components in the proposed framework was further validated through several ablation studies.

{\small
\bibliographystyle{ieee_fullname}
\bibliography{egbib}
}

\end{document}


\title{Robustified Domain Adaptation\\
\normalfont\Large Supplementary Materials}

\author{%
Jiajin Zhang$^*$ \quad Hanqing Chao\thanks{J.Z. and H.C. are co-first authors.} \quad Pingkun Yan\thanks{Corresponding author.}\\
Deep Imaging Analytics Lab (DIAL) \\ Rensselaer Polytechnic Institute \\ \{zhangj41, chaoh, yanp2\}@rpi.edu}

\maketitle

\section{More Implementation Details on DIGITS and Office-31}
For data prepropcessing, we resized the images in DIGITS (MNIST~\cite{LeCun1998MNIST}, USPS~\cite{Hull1994USPS}, MNIST-M~\cite{Ganin2016DANN}), and Office-31~\cite{saenko2010adapting} to $28\times28$ and $224\times224$, respectively, using bilinear interpolation. Following the previous adversarial defense works~\cite{CW2017evaluate,Madry2018Towards,Jordan2019TRADE}, all images are normalized into the range of $[0, 1]$. $l_{\infty}$-norm is used as the metric to measure the bound of adversarial perturbation. Supplementary Tab.~\ref{tab:net} shows the architecture of the modified LeNet on the DIGITS datasets. We used the standard ResNet-50 network with 31 classes output on the Office-31 dataset. All our networks were implemented using PyTorch v1.3.0 and the code will be publicly available through GitHub.

\begin{table}[h]
\centering
\caption{Network architecture of the encoder used on DIGITS datasets.}
\scalebox{0.9}{
\begin{tabular}{c}
\toprule
\textbf{LeNet Architecture}\\
\midrule
Conv(5$\times$5,1,20) + BN2d + Maxpooling(2$\times$2,2,20) + Relu \\
\hline
Conv(5$\times$5,1,50) + BN2d + Maxpooling(2$\times$2,2,20) + Relu \\
\hline
FC(800,100) + BN1d + Relu \\
\hline
FC(100,10) + Relu \\
\bottomrule
\end{tabular}
}
\label{tab:net}
\end{table}

\begin{table*}[t!]
\centering
\caption{The adversarial data robustness (\%) of CURDA on DIGITS. The robust ADDA and robust SRDC represent CURDA-robustified ADDA and SRDC model, respectively.}
\scalebox{0.90}{
\begin{tabular}{c||c||c|c|c}
\toprule
Model & Attack & MNIST$\rightarrow$USPS & USPS$\rightarrow$MNIST & MNIST$\rightarrow$MNIST-M \\
\hline
\hline
\multirow{3}{*}{Robust ADDA} & FGSM & 86.3 & 84.9 & 66.3 \\
\cline{2-5}
& PGD-40 & 77.5 & 76.6 & 60.6 \\
\cline{2-5}
& PGD-100 & 77.4 & 76.4 & 60.3 \\
\hline
\hline
\multirow{3}{*}{Robust SRDC} & FGSM & 93.2 & 94.5 & 82.8 \\
\cline{2-5}
& PGD-40 & 90.1 & 92.2 & 76.8 \\
\cline{2-5}
& PGD-100 & 90.0 & 92.1 & 76.6 \\
\bottomrule
\end{tabular}
}
\label{tab:attack_digits}
\end{table*}

\begin{table*}[t!]
\centering
\caption{The adversarial data robustness(\%) of CURDA on Office-31. The robust ADDA and robust SRDC represent CURDA-robustified ADDA and SRDC model, respectively.}
\scalebox{0.95}{
\begin{tabular}{c||c||c|c|c|c|c|c}
\toprule
Model & Attack & A$\rightarrow$W & W$\rightarrow$D & D$\rightarrow$A & A$\rightarrow$D & D$\rightarrow$W & W$\rightarrow$A \\
\hline
\hline
\multirow{3}{*}{Robust ADDA} & FGSM & 87.8 & 97.5 & 60.8 & 85.4 & 93.9 & 59.4 \\
\cline{2-8}
& PGD-40 & 86.4 & 95.8 & 54.1 & 83.5 & 92.5 & 54.9 \\
\cline{2-8}
& PGD-100 & 86.3 & 95.8 & 53.9 & 83.5 & 92.3 & 54.8 \\
\hline
\hline
\multirow{3}{*}{Robust SRDC} & FGSM & 90.2 & 98.1 & 71.8 & 90.7 & 97.6 & 72.4 \\
\cline{2-8}
& PGD-40 & 89.1 & 96.1 & 66.8 & 88.6 & 95.4 & 67.0 \\
\cline{2-8}
& PGD-100 & 89.0 & 96.0 & 66.7 & 88.5 & 95.3 & 66.8 \\
\bottomrule
\end{tabular}
}
\label{tab:attack_office}
\end{table*}

\section{More Details of UDA Models}

\subsection{Adversarial Discriminative Domain Adaptation (ADDA)}
The ADDA~\cite{tzeng2017adversarial} model includes two separate encoders (\textit{i.e.}, a source encoder $F^{s}$ with fixed parameters and a target encoder $F^{uda}$) and one domain discriminator $D$. In our experiments, both of the two encoders are initialized with the pretrained source robust model $F^{sr}$. The adversarial-discriminative loss in ADDA is used as the original UDA loss $L_{uda}$ in CURDA. The domain discriminator $D$ is optimized alternatively with the target encoder $F^{uda}$. The loss function used to train the domain discriminator $D$ is formulated as
\begin{align}
    \mathcal{L}_{D} = & -\mathbb{E}_{\mathbf{x}^s\sim\mathcal{S}}[\log D(F^{s}_f(\mathbf{x}^s))] \nonumber\\
    & -\mathbb{E}_{\mathbf{x}^t\sim\mathcal{T}}[\log(1-D(F^{uda}_f(\mathbf{x}^t)))].
\end{align}
The target domain encoder $F^{uda}$ is trained by optimizing the UDA loss function of
\begin{equation}
    \mathcal{L}_{uda} = -\mathbb{E}_{\mathbf{x}^t\sim\mathcal{T}}[\log D(F^{uda}_f(\mathbf{x}^t))].
\end{equation}

\subsection{Structurally Regularized Deep Clustering (SRDC)}
The SRDC loss $L_{SRDC}$ proposed in~\cite{tang2020unsupervised} was employed as the original UDA loss $L_{uda}$ in CURDA robustified SRDC.

Let $\{\mathbf{p}_i^t = F^{uda}_c(\mathbf{x}^t_i)\}_{i=1}^{N_t}$ denote the probability prediction of the UDA model on the target domain. Denote the probability of the $k^{th}$ class as $p_{i,k}^t, k\in\{1,...,M\}$. A set of auxiliary probability vectors $\{\mathbf{q}_i^t\}_{i=1}^{N_t}$ are defined as
\begin{equation}
    q_{i,k}^t = \frac{p_{i,k}^t/(\sum_{i'=1}^{N_t} p_{i',k}^t)^{\frac{1}{2}}}{\sum_{k'=1}^{M} p_{i,k'}^t/(\sum_{i'=1}^{N_t} p_{i',k'}^t)^{\frac{1}{2}}}.
\end{equation}
We use $P^t$ and $Q^t$ to represent the distribution of $\{\mathbf{p}_i^t\}_{i=1}^{N_t}$ and $\{\mathbf{q}_i^t\}_{i=1}^{N_t}$. The first term of the SRDC loss is given as
\begin{equation}
\label{eq:fphi_t}
    \mathcal{L}_{c}^t = \textbf{\textit{KL}}(Q^t||P^t) + \sum_{k=1}^M\rho_k^t log \rho_k^t,
\end{equation}
where $\rho_k^t = \frac{1}{N_t}\sum_{i=1}^{N_t} q_{i,k}^t$ and $\textbf{\textit{KL}}(\cdot)$ represents the KL-Divergence.

Let $\{\boldsymbol{\mu}_k\}_{k=1}^{M}$ be the learnable cluster centers of both the source and target data in the feature space (for more details, please refer to~\cite{tang2020unsupervised}). The authors further introduce a set of probability vectors $\{\Tilde{\mathbf{p}}_i^t\}_{i=1}^{N_t}$ defined as 
\begin{equation}
    \Tilde{\mathbf{p}}_{i,k}^t = \frac{exp((1+||F_f^{uda}(\mathbf{x}_i^t)-\boldsymbol{\mu}_k||^2)^{-1})}{\sum_{k'=1}^M exp((1+||F_f^{uda}(\mathbf{x}_i^t)-\boldsymbol{\mu}_k'||^2)^{-1})}.
\end{equation}

Let ${\Tilde{P}^t}$ denote the distribution of $\{\Tilde{\mathbf{p}}_i^t\}_{i=1}^{N_t}$. 
${\Tilde{Q}^t}$ is then introduced as the counterpart of ${\Tilde{P}^t}$ to describe another set of auxiliary probability vectors $\{\Tilde{\mathbf{q}}_i^t\}_{i=1}^{N_t}$. Here, $\{\Tilde{\mathbf{q}}_i^t\}_{i=1}^{N_t}$ is defined based on $\{\Tilde{\mathbf{p}}_i^t\}_{i=1}^{N_t}$ similarly with $\{\mathbf{q}_i^t\}_{i=1}^{N_t}$.
We then have the second term of the SRDC loss:
\begin{equation}
\label{eq:phi_t}
    \mathcal{L}_{f}^t = \textbf{\textit{KL}}(\Tilde{Q}^t||\Tilde{P}^t) + \sum_{k=1}^M \Tilde{\rho}_k^t log \Tilde{\rho}_k^t,
\end{equation}
where $\Tilde{\rho}_k^t = \frac{1}{N_t}\sum_{i=1}^{N_t} \Tilde{q}_{i,k}^t$.

The target domain SRDC loss $\mathcal{L}_{SRDC}^t$ is given as a joint loss of Supplementary Equ.~\ref{eq:fphi_t} and \ref{eq:phi_t}:
\begin{equation}
    \mathcal{L}_{SRDC}^t = \mathcal{L}_c^t + \mathcal{L}_f^t.
\end{equation}

Similarly, the source domain SRDC loss $\mathcal{L}_{SRDC}^s$ is computed by replacing the the $\{\mathbf{q}_i^t\}_{i=1}^{N_t}$ with the one-hot indicator vectors of source labels.
The final original UDA loss function $\mathcal{L}_{uda}$ is formulated as
\begin{equation}
\label{eq:srdc}
    \mathcal{L}_{uda} = \mathcal{L}_{SRDC}^t + \lambda_{p}\mathcal{L}_{SRDC}^s,
\end{equation}
where $\lambda_{p}$ is a positive weighting parameter. The hyperparameter settings used in our experiments are the same as the original paper.

\section{Ablation Studies on the CURDA Robustified SRDC}

Supplementary Tab.~\ref{tab:Ablat_supp} shows the results of ablation studies on the CURDA robustified SRDC. Similar with the experiments in Sec.~4.5, we performed a leave-one-component-out study of each component on two tasks of Office-31 (A$\rightarrow$W  and  W$\rightarrow$A).

\begin{table}[ht]
\centering
\caption{The results of ablation Studies on the CURDA robustified SRDC. Rows 2-4 present the effects of three major losses.
The row '\textit{w/o. switch lab-gen}' means pseudo label generation without switching encoders.}
\scalebox{0.87}{
\begin{tabular}{l||c|c|c|c}
\toprule
\multirow{2}{*}{Methods} & \multicolumn{2}{c|}{A$\rightarrow$W} & 
\multicolumn{2}{c}{W$\rightarrow$A}\\
\cline{2-5}
& Clean & Rob. & Clean & Rob. \\
\hline
\hline
w/o. $\mathcal{L}_{con}$ & 86.3$\pm$0.1 & 80.8$\pm$0.2 & 61.9$\pm$0.2 & 54.7$\pm$0.2\\
\hline
w/o. $\mathcal{L}_{saa}$ & 49.0$\pm$0.3 & 41.5$\pm$0.2 & 38.6$\pm$0.3 & 30.1$\pm$0.3\\
\hline
w/o. $\mathcal{L}_{trade}$ & 88.6$\pm$0.1 & 87.2$\pm$0.2 & 69.7$\pm$0.2 & 63.4$\pm$0.2\\
\hline
w/o. switch & \multirow{2}{*}{83.3$\pm$0.2} & \multirow{2}{*}{78.6$\pm$0.2} & 
\multirow{2}{*}{66.7$\pm$0.3} & \multirow{2}{*}{60.2$\pm$0.2}\\
label gen. &  &  &  & \\
\hline
CURDA & {\bf90.5}$\pm$0.2 & {\bf89.1}$\pm$0.3 & {\bf72.8}$\pm$0.2 & {\bf67.0}$\pm$0.2\\
\bottomrule
\end{tabular}
}
\label{tab:Ablat_supp}
\end{table}

\section{Results Under Different Adversarial Attacks}
As mentioned in Sec.~4.3, for datasets containing binary images, i.e. MNIST and USPS, the attack perturbation and perturbation step size were set to 0.3 and 0.01, respectively. For colorful image datasets (MNIST-M and Office-31), the two parameters were set to 0.031 and 0.007, respectively. 
In this section, we verified the convergence of the Projected Gradient Descent (PGD) \cite{Madry2018Towards} attack by changing the number of attack steps. The model robustness was examed under 40- and 100-steps PGD attack, which are denoted as ``PGD-40'' and ``PGD-100'', respectively. In addition, we also report the results with another standard attack method, Fast Gradient Sign Method (FGSM) \cite{Goodfellow2014advsamples}.
The results are shown in Supplementary Tab.~\ref{tab:attack_digits}\&\ref{tab:attack_office}. It can be seen that the adversarial robustness of CURDA robustified models (both ADDA and SRDC) changed less than $0.3\%$, when increasing the perturbation steps from 40 to 100. This indicates the convergence of PGD attacks.

\section{Results on VisDA and CIFAR-STL}

We used ResNet-101~\cite{he2016deep} and WideResNet(WRN-34-10)~\cite{Zagoruyko2016WRNet} as the backbone encoders to further evaluate the CURDA on CIFAR-STL~\cite{Krizhevsky2009Cifar,coates2011analysis} and VisDA-2017~\cite{peng2017visda}, respectively. 
Stochastic gradient descent (SGD) with a momentum of 0.9 was used as the optimizer to train all the networks. For each experiment, we trained the model from scratch three times with different random seeds and reported the averaged performance in the form of mean($\pm$std).

\subsection{VisDA}
\textbf{VisDA-2017}~\cite{peng2017visda} benchmark is a UDA challenge focusing on the simulation-to-reality adaptation. There are over 280,000 images across 12 categories. We used the $152,397$ synthetic images as the source domain, and the $55,388$ real images as the target domain. In the target domain, we randomly select $40,000$ images for training and the rest $15,388$ real images were preserved for testing.

On VisDA-2017, the source-robust model (ResNet-101) was trained for $40$ epochs with a learning rate of $0.001$ in the first $15$ epochs and with a learning rate of $0.0001$ for the rest of the epochs. The UDA model is trained for $40$ epochs with a learning rate of $0.001$. A weight decay of $2e-4$ and a batch size of $64$ is used in both source robust model training and UDA. The pseudo label generator uses $\tau=15$ and $P_{pseudo}=0.90$. The margins $m_s$ and $m_t$ are all set to be 20. In the loss functions, we set $\lambda_{con}=\lambda_{saa}=0.01$.

The comparison results of CURDA and the other three baseline models is presented in Supplementary Tab.~\ref{tab:visda}. In  comparison,  only  the  CURDA  obtained  steady high improvement on robustness and also maintained a competitive accuracy on clean data.

\begin{table}[t]
	\centering
\caption{Clean data accuracy (Clean (\%)) and adversarial data robustness (Rob. (\%)) on the VisDA-2017 benchmark. The best results on each task are shown in \textbf{bold}.}
	\label{tab:visda}
	\scalebox{0.80}{
	\begin{tabular}{l|c||c|c}
		\toprule
		\multirow{2}{*}{\textbf{UDA models}} & \multirow{2}{*}{\textbf{AT methods}} & \multicolumn{2}{c}{\textbf{Syn.$\rightarrow$Real}}\\
		\cline{3-4}
		& & Clean & Rob.\\
		\hline
		\hline
		\multirow{3}{*}{\textbf{ADDA}} & None & {\bf65.7}$\pm$0.2 & 0.0$\pm$0.0 \\ 
		\cline{2-4}
		& AT in SD & 62.2$\pm$0.2 & 2.8$\pm$0.1 \\ 
		\cline{2-4}
		& AT in SD \& TD & 35.5$\pm$0.2 & 12.9$\pm$0.3 \\
		\cline{2-4}
		& CURDA & 61.4$\pm$0.2 & {\bf51.7}$\pm$0.2\\ 
		\hline
		\hline
		\multirow{3}{*}{\textbf{SRDC}} & None & {\bf83.9}$\pm$0.1 & 0.0$\pm$0.0 \\ 
		\cline{2-4}
		& AT in SD & 78.7$\pm$0.2 & 23.6$\pm$0.2 \\ 
		\cline{2-4}
		& AT in SD \& TD & 70.1$\pm$0.2 & 59.2$\pm$0.2 \\
		\cline{2-4}
		& CURDA & 78.0$\pm$0.2 & {\bf69.4}$\pm$0.2\\ 
		\bottomrule
	\end{tabular}
	}
\end{table}

\subsection{CIFAR-STL}
CIFAR-STL is composed by two colour image classification datasets, CIFAR-10 ~\cite{Krizhevsky2009Cifar} with 60,000 images and STL-10~\cite{coates2011analysis} with 1,300 images. CIFAR-10 is also widely used for evaluating model robustness against adversarial attack. In out expriments, the CIFAR-10 was used as the source domain and the STL-10 is used as the target domain. On the STL-10, $900$ images were randomly selected as the training target set and the rest $400$ images were preserved as the test set.
Both STL-10 and CIFAR-10 are 10-class image datasets. These two datasets contain 9 overlapping classes. To focus on evaluating UDA, we removed the non-overlapping classes, `frog' and `monkey' from the two datasets. 

On CIFAR-STL, we trained the source-robust model (WRN-34-10) for $80$ epochs with a learning rate of $0.001$ in the first $55$ epochs and changed the learning rate to $0.0005$ for the rest epochs. The target encoder is trained for $70$ epochs on a learning rate of $0.001$. A weight decay of $2e-4$ and a batch size of $128$ is used in both source training and adaptation. The pseudo label generator uses $\tau=30$ and $P_{pseudo}=0.90$. The margins $m_s$ and $m_t$ are all set to be 8. In the loss functions, we set $\lambda_{con}=\lambda_{saa}=0.01$.

The results of the CURDA and the three baseline models are presented in Supplementary Tab.~\ref{tab:cifar}. We can observe that the proposed CURDA obtained the highest UDA model robustness with small accuracy drop on clean samples.

\begin{table}[t]
	\centering
\caption{Clean data accuracy (Clean (\%)) and adversarial data robustness (Rob. (\%)) on the CIFAR-STL benchmark. The best results on each task are shown in \textbf{bold}.}
	\label{tab:cifar}
	\scalebox{0.80}{
	\begin{tabular}{l|c||c|c}
		\toprule
		\multirow{2}{*}{\textbf{UDA models}} & \multirow{2}{*}{\textbf{AT methods}} & \multicolumn{2}{c}{\textbf{CIFAR$\rightarrow$STL}}\\
		\cline{3-4}
		& & Clean & Rob.\\
		\hline
		\hline
		\multirow{3}{*}{\textbf{ADDA}} & None & {\bf77.6}$\pm$0.2 & 0.0$\pm$0.0 \\ 
		\cline{2-4}
		& AT in SD & 76.4$\pm$0.2 & 10.4$\pm$0.1 \\ 
		\cline{2-4}
		& AT in SD \& TD & 45.0$\pm$0.2 & 22.2$\pm$0.2 \\
		\cline{2-4}
		& CURDA & 74.2$\pm$0.2 & {\bf47.9}$\pm$0.2\\ 
		\hline
		\hline
		\multirow{3}{*}{\textbf{SRDC}} & None & {\bf78.6}$\pm$0.2 & 0.0$\pm$0.0 \\ 
		\cline{2-4}
		& AT in SD & 77.1$\pm$0.2 & 11.6$\pm$0.1 \\ 
		\cline{2-4}
		& AT in SD \& TD & 45.6$\pm$0.2 & 23.3$\pm$0.2 \\
		\cline{2-4}
		& CURDA & 75.9$\pm$0.2 & {\bf48.1}$\pm$0.2\\ 
		\bottomrule
	\end{tabular}
	}
\end{table}

\bibliographystyle{ieee_fullname}
\bibliography{egbib}